\documentclass[10pt,journal,compsoc]{IEEEtran}
\usepackage{graphicx}
\usepackage{amsmath}
\usepackage{amssymb}
\usepackage{booktabs}
\usepackage {xcolor}

\ifCLASSOPTIONcompsoc
  \usepackage[nocompress]{cite}
\else
  \usepackage{cite}
\fi

\ifCLASSINFOpdf
\else
\fi

\begin{document}
%
\title{Hierarchical Attention Network for Few-Shot Object Detection via Meta-Contrastive Learning}

\author{Dongwoo Park,~Jong-Min Lee
\IEEEcompsocitemizethanks{\IEEEcompsocthanksitem D. Park is with the Department of Artificial Intelligence. E-mail: infinity7428@hanyang.ac.kr\protect
\IEEEcompsocthanksitem J. Lee is with the Department of Biomedical Engineering. E-mail: ljm@hanyang.ac.kr\protect

\IEEEcompsocthanksitem D. Park and J. Lee are with Hanyang University.}
\thanks{Manuscript received November 30, 2022;}}

\markboth{Journal of \LaTeX\ Class Files,~Vol.~14, No.~8, August~2015}%
{Shell \MakeLowercase{\textit{et al.}}: Bare Demo of IEEEtran.cls for Computer Society Journals}

\IEEEtitleabstractindextext{%
\begin{abstract}
Few-shot object detection (FSOD) aims to detect objects of a novel category in an image with only a few instances.
Recently, research has revealed that metric-based meta-learning methods have high adaptability and exhibits good performance.
However, existing FSOD methods still have drawbacks.
(1) Existing meta-learning based FSOD methods focus on global attention during the feature aggregation of the query and support images. However, these methods do not fully utilize the local context information of an image. Therefore, we propose a novel hierarchical attention module (HAM) that hierarchically extends receptive fields from local to global.
(2) Existing studies have not been investigated whether meta-learning works well in FSOD. Does using meta-learning to solve FSOD problems really work?
In this study, we explore whether the meta-learning approach works well in FSOD and propose a meta-contrastive learning module based on correlation similarity to help the objective function of meta-learning to perform well. 
Finally, we establish a new state-of-the-art network, by realizing significant margins. Our code is available at: https://github.com/infinity7428/hANMCL

\end{abstract}

}

\maketitle

\IEEEdisplaynontitleabstractindextext

\IEEEpeerreviewmaketitle

\IEEEraisesectionheading{\section{Introduction}\label{sec:introduction}}

\IEEEPARstart{I}{n} recent years, object detection has undergone tremendous development, and excellent performance detectors have been proposed \cite{ren2015faster, lin2017feature,  he2017mask, cai2018cascade}. Object detection requires a large number of annotated images and considerable time, labor, and cost. In low-data scenarios, showing good performance is difficult with only a small amount of data owing to diversity. Therefore, few-shot object detection (FSOD) has attracted considerable attention because it does not require a large amount of data.
Significant research has been conducted on meta-learning approaches to solve FSOD problems.
Meta-learning methods use support images to extract and aggregate helpful features for evaluating query images.
A key element of meta-learning based FSOD is how features are aggregated.
However, despite extensive research in recent years, existing FSOD methods still have two major drawbacks.

\begin{figure*}[t]
\centering
\includegraphics[width=0.9\textwidth]{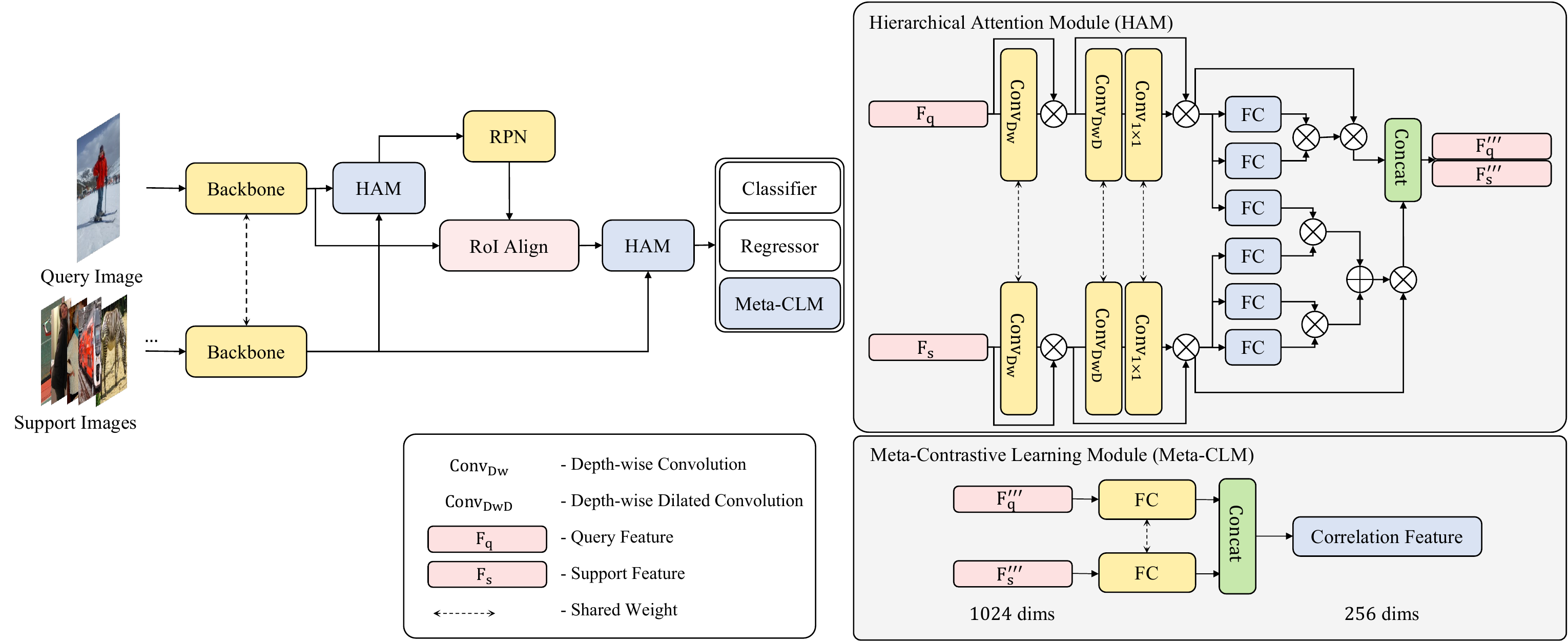}
\caption{Framework of the proposed architecture. Query and support images are processed by the hierarchical attention module (HAM), and are then efficiently exploited through global and cross attention. $\text{Conv}_\text{{Dw}}$: depth-wise convolution; $\text{Conv}_\text{{DwD}}$: depth-wise dilated convolution. Meta-CLM generates correlation feature by passing query and support features through the single fully connected layer it shares.}
\label{fig1}
\end{figure*}

(1) We tackle the structural limitations of feature aggregation methods.
Existing methods \cite{hsieh2019one, hu2021dense, chen2021dual, yan2019meta} mainly use the global attention in the process of extracting and aggregating features from the query and support image.
However, this global attention mechanism ignores important local context information in the image, so local context information is not fully utilized. 
In previous studies \cite{raghu2021vision, park2022vision}, research has been conducted that convolution-based and global attention-based blocks have different pros and cons, and that they can complement each other.
We point out the problem of global attention, which is mainly used in the feature aggregation method, and propose a method combining it with convolution-based attention to compensate for the shortcomings of global attention.
As Raghu et al. \cite{raghu2021vision} explained, when comparing feature maps sequentially extracted from lower and higher layers, the network using only global attention shows a high feature representation similarity. On the other hand, in the convolution-based network, the similarity between lower and higher layers were low. This means that there is a clear difference in the way convolution-based and global attention-based networks generate feature maps.
As Park et al. \cite{park2022vision} explained, global attention-based network is robust against high-frequency noises. But convolution-based network is robust against low-frequency noises.
When high frequency-based random noises were added as input of convolution-based network, the performance decrease was large.
Convolution-based network is vulnerable to high-frequency noise, while global attention-based network is robust against them.
In the evaluation of the convolution only network, the removal of the lower layer showed a serious performance decrease, and in the global case, the removal of the higher layer showed a serious performance decrease.
Based on the above observations, we find how to harmonize the pros and cons of convolution-based and global attention-based networks.
So, we propose a feature aggregation module that combines the advantages of convolution-based attention and global attention, called HAM.

(2) We tackle how the objective function of meta-learning works.
In general, researchers use a large number of meta-learning approaches to solve few-shot learning problems. However, it is necessary to investigate \emph{whether this actually works.}
Meta-learning is a method that helps evaluate query images well by extracting and aggregating helpful features using support images.
The commonly employed aggregation methods include Vinyals et al. \cite{vinyals2016matching} and Sung et al. \cite{sung2018learning}.
In many studies \cite{hsieh2019one, xiao2020few, hu2021dense, zhang2021accurate, zhang2021meta, chen2021dual, li2021transformation} using meta-learning, better classification is achieved by simply combining support images rather than learning them based on similarity. This aggregate and classify query and support images to create a feature space that can be distinguished by category.
However, only general classification can be achieved, given the support image, and it is not the most appropriate use of meta-learning.
Therefore, unlike other studies, we changed multi-class classification to binary classification in this study to measure the similarity between the query image and the support image, regardless of the category.
We aim to create a class-agnostic feature space and measure the similarity between the query image and the support image. In addition, we propose a meta-contrastive learning module that can better compare similarities between query and support images by generating correlation features.
The correlation feature comprises combinations of query features and support features.
This method constructs a feature space that distinguishes between query and support images that are concatenated in the same class and those that are not.
The model can extract features that better consider the correlation between two images regardless of the category.
Therefore, this strategy improves the performance of meta-learning based FSOD.

Our proposed method is a two-stage detector based on Faster R-CNN ResNet-101. This structure is composed of a hierarchical attention module (HAM) and meta-contrastive learning module (Meta-CLM).
We achieved significant performance improvements regardless of the number of images on the COCO dataset \cite{lin2014microsoft}. In particular, the proposed method established a new state-of-the-art method with large margins. Our proposed method brings 0.3, 1.4, 1.4, 1.6, 2.4, and 3.2\% AP improvements for 1-30 shots object detection on COCO dataset. The main contributions of our approach are as follows.

\begin{itemize}
    \item We propose a feature aggregation module that extends the receptive field called HAM that can compensate for the shortcomings of global attention.
    \item We propose a novel contrastive learning module that enhances the similarity between the query image and support image, and matches it with the objective function of meta-learning.
    \item We establish a new state-of-the-art network, by realizing significant margins.
    \item Our proposed FSOD network can strongly detect novel categories without a fine-tuning stage in 1-30 shots object detection on COCO dataset. 
\end{itemize}

\begin{figure*}[t]
\centering
\includegraphics[width=.9\textwidth]{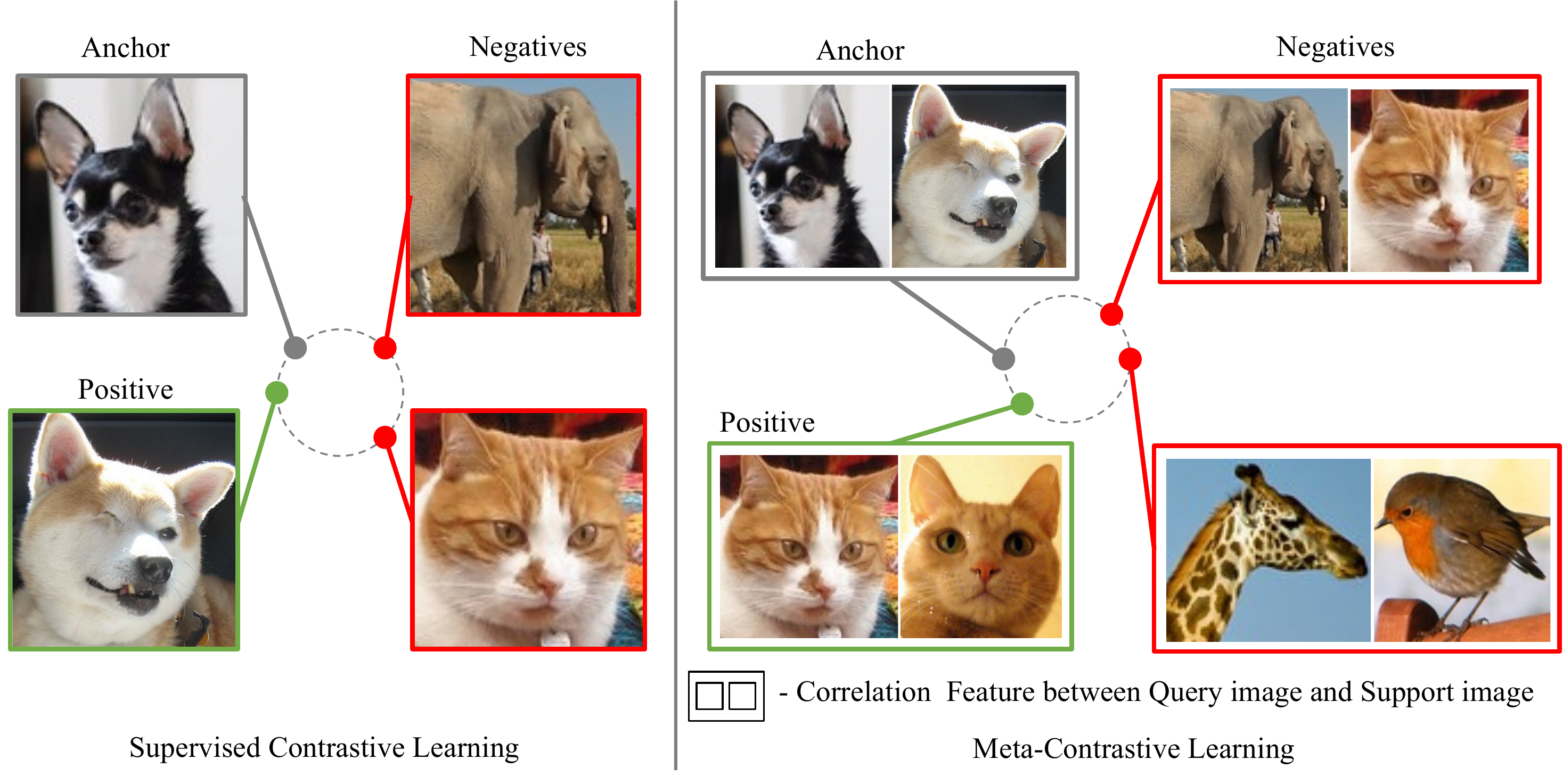} 
\caption{Conceptual visualization of supervised contrastive learning vs. meta-contrastive learning. (Left) contrasts the set of all samples from the same class as positives against the negatives from the remainder of the batch. (Right) correlation feature between query and support images is a basic unit. Meta-contrastive learning contrasts the set of correlation features from the proposals of the same image as \textit{positives}.}
\label{fig2}
\end{figure*}

\section{Related Work}
\label{sec:formatting}
\subsection{Few-Shot Object Detection}
FSOD can be divided into meta-learning \cite{kang2019few, hsieh2019one, yan2019meta, fan2020few} and transfer learning methods \cite{qiao2021defrcn, cao2021few, sun2021fsce}.
(1) Meta-learning solves the problem of matching a support image to the query image when the support and query images are given; this is a method that has been widely used in existing few-shot classification approaches.
Early FSOD methods used a meta-learning strategy. Kang et al. \cite{kang2019few} proposed a meta-learning method based on YOLO v2 that extracts the global features of the support image using a re-weighting module. It then acquires the global features of the support image using the re-weighting module and uses them as a coefficient to refine the query image.
Inspired by the squeeze-and-excitation method, Hsieh et al. \cite{hsieh2019one} proposed co-excitation to make the query features attend to the support features, and vice versa.
Yan et al. \cite{yan2019meta} demonstrated a method to aggregate support and query features through channel-wise multiplication using a meta-learner.
Fan et al. \cite{fan2020few} proposed a module that aggregates support and query images in an RPN to extract features that distinguish images from other categories.
(2) Transfer learning methods refer to reusing the weights of a network trained with abundant $D_{base}$.
First, the model is pre-trained using abundant data from base categories $D_{base}$ and solves a few-shot learning problem using weights through the fine-tuning step with novel category images.
Qiao et al. \cite{qiao2021defrcn} stopped the gradient in the RPN because the RPN must generate many proposals regardless of the class, and the region of interest (RoI) head performs class-specific classification and regression.
Cao et al. \cite{cao2021few} associated and discriminated the data from novel categories to base categories to ensure inter-class separability and intra-class compactness.
Sun et al. \cite{sun2021fsce} proposed a fine-tuning method that adds a contrastive head to the RoI head to perform contrastive learning between proposals.

\subsection{Attention Mechanisms}
FSOD using an attention mechanism is intended to high-light relevant features. Previous studies have been conducted on extracting features from query and support images using an attention mechanism with a global receptive field \emph{(i.e., self-attention, cross-attention, non-local block, etc.)}
Sung et al. \cite{sung2018learning}-based methods \cite{hu2021dense, chen2021dual} have been previously developed. However, they still use only global attention for the feature aggregation, which has structural limitations, and the features of support images are not fully exploited.
Hu et al. \cite{hu2021dense} uses only cross-attention between query and support images as a way to utilize support images.
Furthermore, Chen et al. \cite{chen2021dual} attempts to capture the relationship between channels of the support image.

However, local context information is still not utilized and sufficient feature extraction for query and support images is not achieved.
So we propose a method to compensate for the shortcomings of existing feature aggregation modules.

\begin{table*}[t]
\centering
\caption{Performance on Pascal VOC dataset (1,2,3,5, and 10 shots). Avg.: result of averaging 10 multiple runs. Way: category of classification -: no result in the paper.}
\includegraphics[width=0.9\textwidth]{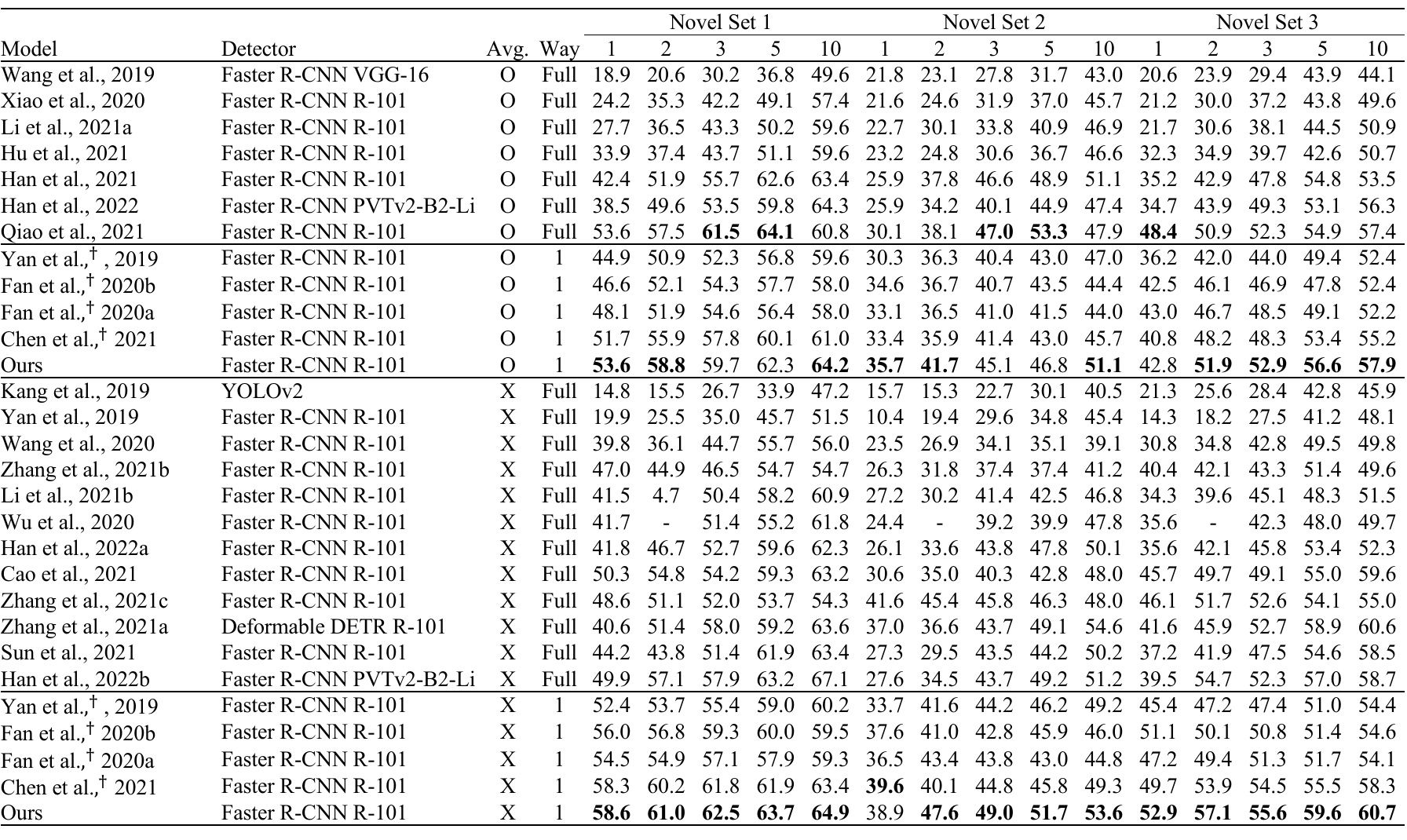} 
\label{table1}
\end{table*}

\begin{table*}[t]
\centering
\caption{Performance on the MS COCO dataset (1,2,3, and 5 shots). † indicates re-implemented results. nAP: novel categories average precision (AP).}
\includegraphics[width=0.75\textwidth]{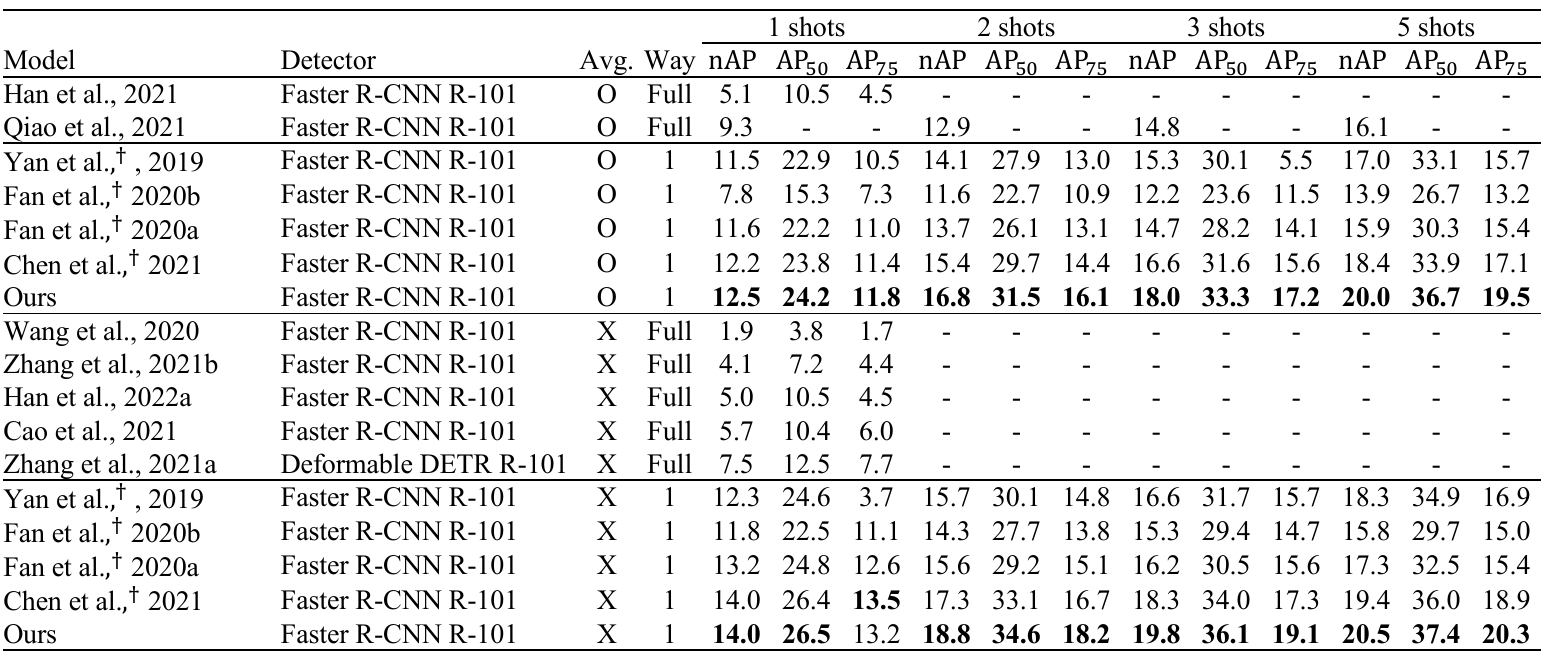} 
\label{table2}
\end{table*}

\begin{table*}[t]
\centering
\caption{Performance on the MS COCO dataset (10 shots).}
\includegraphics[width=0.75\textwidth]{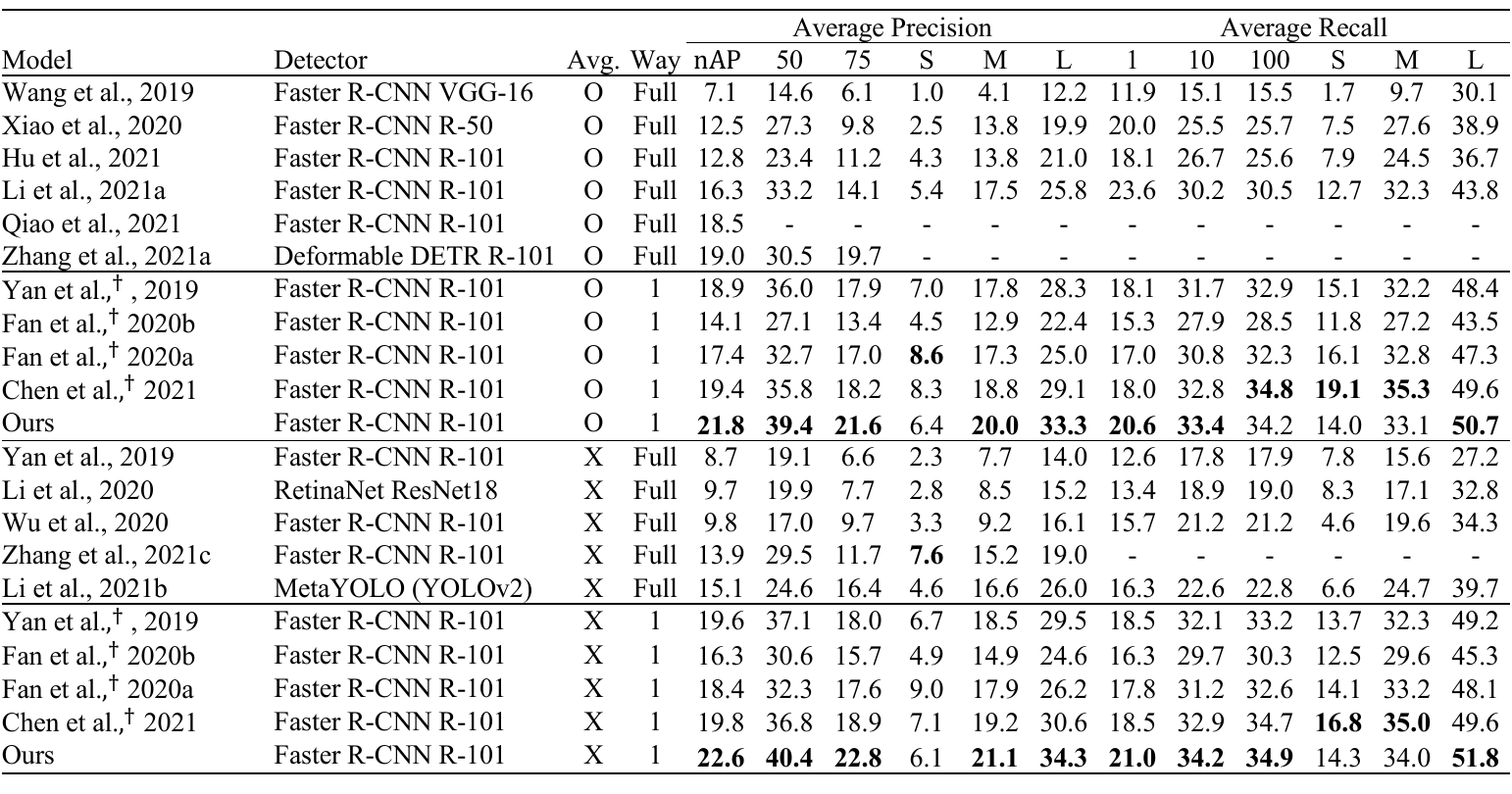} 
\label{table3}
\end{table*}

\begin{table*}[hbt!]
\centering
\caption{Performance on the MS COCO dataset (30 shots).}
\includegraphics[width=0.75\textwidth]{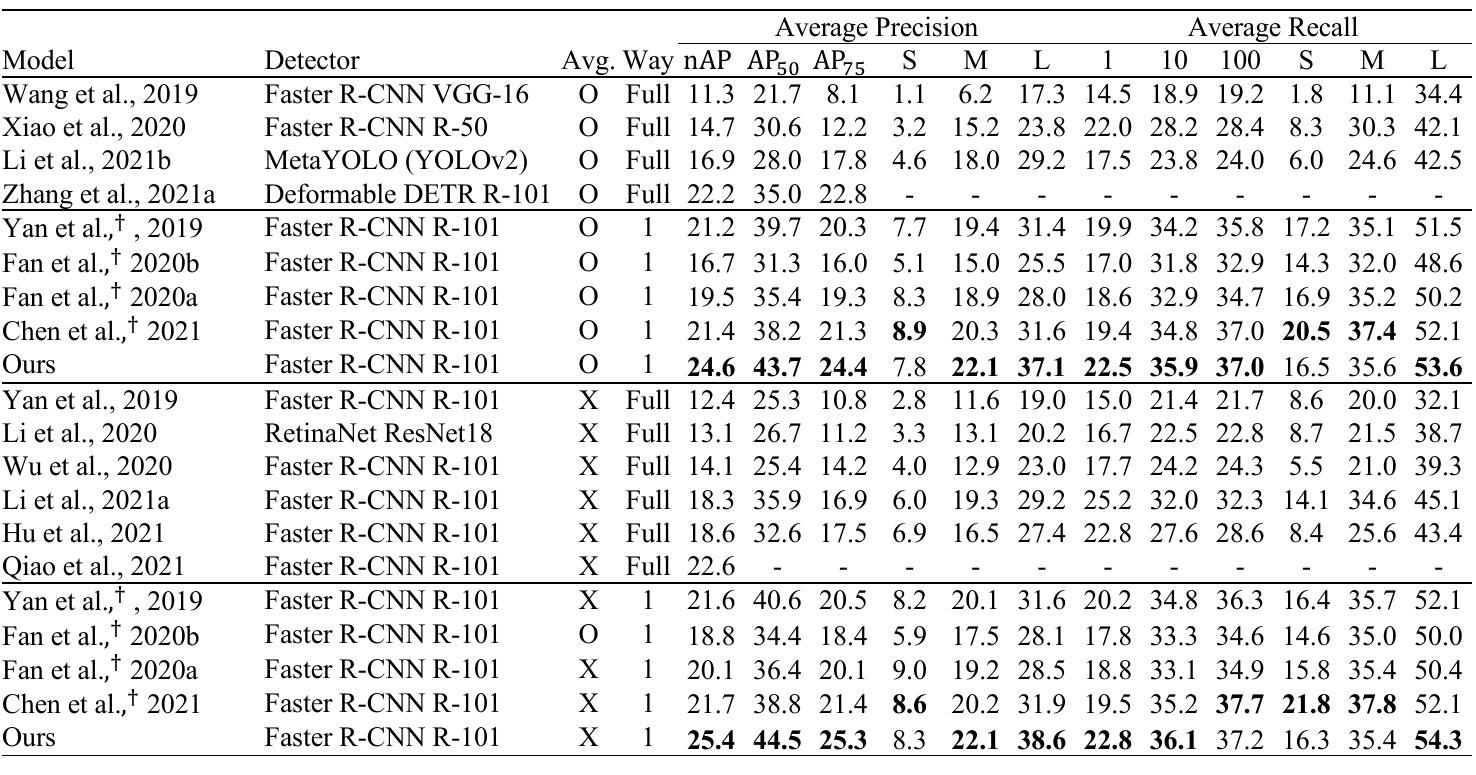} 
\label{table4}
\end{table*}

\section{Method}
First, we review the preliminaries of the FSOD setting. Then, we introduce our method, which addresses the proposed hierarchical attention and meta-contrastive learning modules for FSOD.

\subsection{Problem Definition}
For the FSOD dataset, \{x,y\} = $D$ denotes that it contains image x and ground truth label y. They can be divided into two categories:  $D_{novel}$  and $D_{base}$. Note that $C_{base}$ and $C_{novel}$ are disjointed, i.e., $C_{base}$ $\cap$ $C_{novel}$ = $\emptyset$. Each training data consists of an image x and corresponding label y, where y consists of a class label and bounding box. The fundamental purpose of FSOD is to evaluate the data of $D_{novel}$ from only a low shot of $D_{novel}$ using a model trained with abundant data $D_{base}$. Similar to Kang et al. \cite{kang2019few}, we trained using up to 30 or fewer instances (1, 2, 3, 5, 10, and 30) because only K box instances should be used for training or evaluation.
The overall procedure follows the standard transfer learning of Equation 1.
However, the zero-shot evaluation result is attached to evaluate the adaptability.
This is an evaluation of $D_{novel}$, which has never been learned by training, as $D_{base}$ without a transfer learning process.

\begin{equation}  
\mathcal{M}_{init} \overset{D_{base}}{\longrightarrow} \mathcal{M}_{base} \overset{D_{novel}}{\longrightarrow} \mathcal{M}_{novel}
\end{equation}


\subsection{Revisiting Feature Aggregation Module}

The two-stage detector for FSOD can be divided into four main parts: the backbone, the region proposal network (RPN), RCNN, and the feature aggregation module. 
The backbone extracts the feature from query and support images, the RPN generates proposals based on the similarity between the query and support features, and the RCNN head performs classification and refined bounding box regression.
The feature aggregation module aggregates the query and support features, before the RPN and RCNN, to fully exploit them for assisting the RPN and RCNN.
The entire training procedure is illustrated in Fig. 1.
The input image is fed to the backbone to generate a feature map and conveyed sequentially to the RPN and RCNN.
We propose a method for compensating the structural limitations of this module.
This is achieved by adding a convolution-based attention block to the local receptive fields. Furthermore, we construct a hierarchical attention module with a sequentially larger receptive field. The detailed architecture of the module is presented in Section 3.3.

\begin{figure*}[t]
\centering
\includegraphics[width=.85\textwidth]{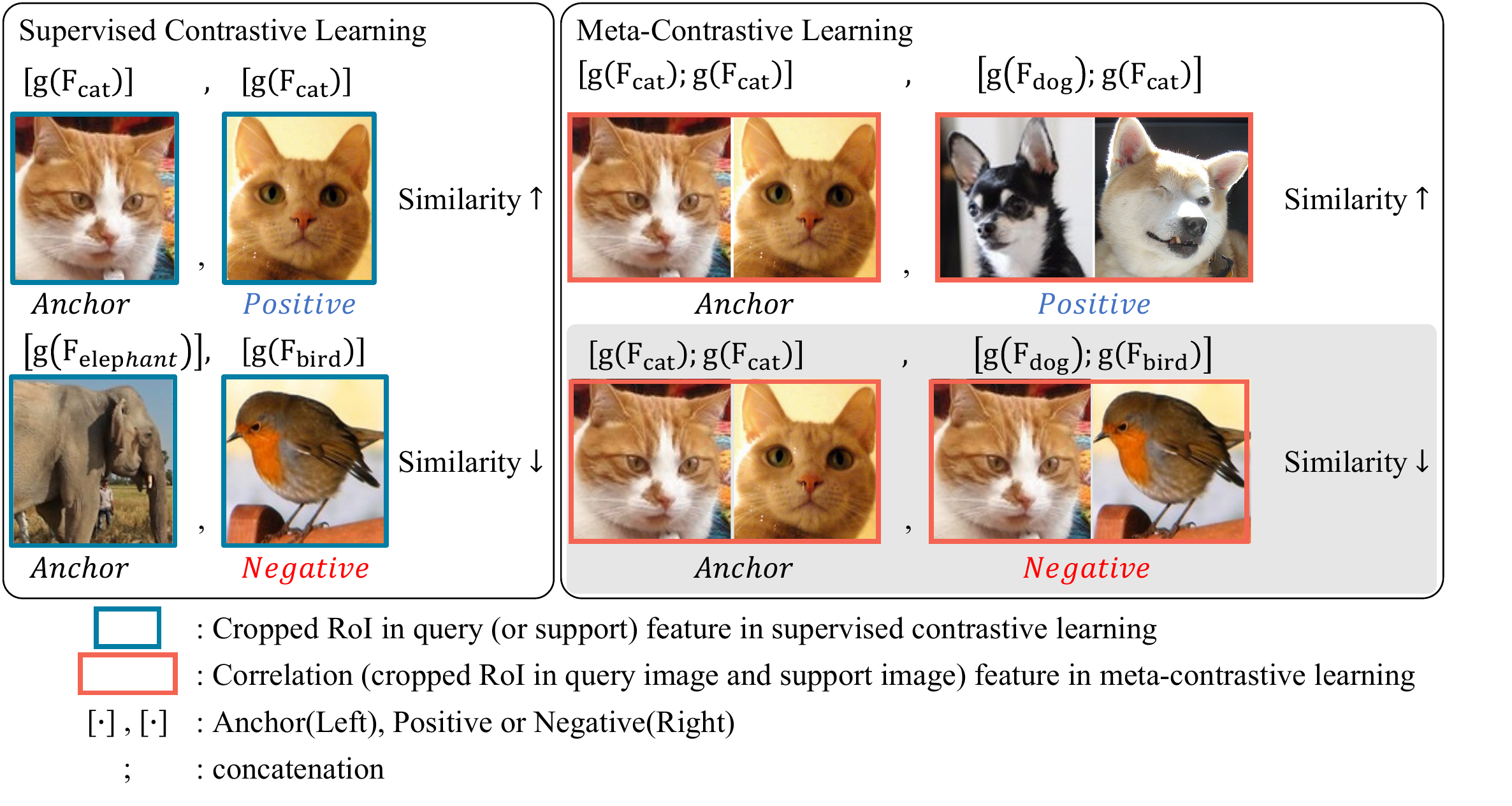} 
\caption{Toy example of Meta-CLM. In meta-contrastive learning module, the query and support features encoded by g($\cdot$) are concatenated. When a correlation feature includes a support image of the same class as the query, and another includes a support image of a different class from the query, the two correlation features are trained to have a lower similarity.}
\label{fig4}
\end{figure*}

\subsection{Hierarchical Attention Module}

\begin{figure}[hbt!]
\centering
\includegraphics[width=.95\columnwidth]{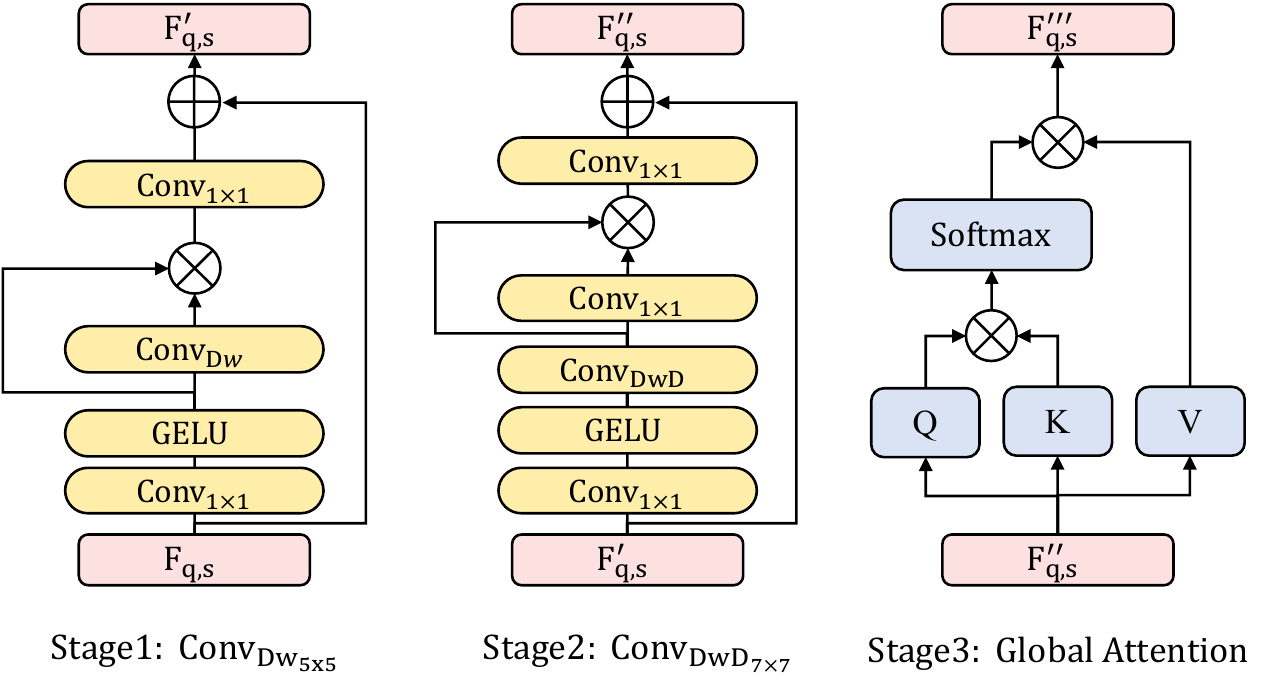} 
\caption{Detailed architecture of the hierarchical attention module (HAM). It consists of three-stage two convolution-based attention and one global attention. Each convolution-based attention layer has a skip-connection.}
\label{fig5}
\end{figure}

\begin{equation} 
F_{q,s}^{'}=\text{Conv}_{\text{Dw}}(F_{q,s})\cdot F_{q,s} + F_{q,s}
\end{equation} 
\begin{equation}  
F_{q,s}^{''}=\text{Conv}_{1\times1}(\text{Conv}_{\text{DwD}}(F_{q,s}^{'}))\cdot F_{q,s}^{'} +  F_{q,s}^{'}
\end{equation}
\begin{equation}  
f(F_{q,s}^{''},F_{q,s}^{''}) = \sigma(\phi{(F_{q,s}^{''})}^{\intercal}\phi^{'}{(F_{q,s}^{''})})
\end{equation}
\begin{equation}  
f(F_{q}^{''},F_{s}^{''}) = \sigma(\phi^{''}{(F_{q}^{''})}^{\intercal}\phi^{'''}{(F_{s}^{''})})
\end{equation}
\begin{equation}  
f(F_{q}^{''},F_{s}^{''})^{'} = f(F_{s}^{''},F_{q}^{''}) + \cdot f(F_{s}^{''},F_{s}^{''})
\end{equation} 
\begin{equation}  
y=[f(F_{q}^{''},F_{s}^{''})\cdot F^{''}_{q} ; f(F_{q}^{''},F_{s}^{''})^{'}\cdot F^{''}_{s}]
\end{equation}  

We point out the limitations of global attention, which is mainly used in the feature aggregation method, and propose a method combining it with a convolution-based attention to compensate for below shortcomings.
As Raghu et al. \cite{raghu2021vision} explained, when comparing feature maps sequentially extracted from lower and higher layers, the network using only global attention shows a high feature representation similarity.
On the other hand, in the convolution-based network, the similarity between lower and higher layers were low. This means that the feature maps extracted by convolution-based and global attention-based networks have different characteristics.
As Park et al. \cite{park2022vision} explained, convolution-based network is vulnerable to high-frequency noise, whereas global attention is robust to them.
In the evaluation of the convolution only network, the removal of the lower layer showed a serious performance decrease, and in the global case, the removal of the higher layer showed a serious performance decrease.
Motivated by the above observations, we have found that the pros and cons of convolution-based and global attention-based networks are opposite to each other.
So, we consider the pros and cons, order, number, and role to design a feature aggregation module that combines the strengths.
The proposed HAM consists of three stages, each with two convolution-based attentions, and one global attention. 
Convolution-based attention contains three types of layers (i.e., $\text{Conv}_\text{{Dw}}$, $\text{Conv}_\text{{DwD}}$,  and $\text{Conv}_{\text{1} \times \text{1}}$, see Fig. 3.).
$\text{Conv}_{\text{Dw}}$ has a local receptive field and helps extract powerful feature representations from an image. The adjacent region, which contains the local context information of the image, utilizes structural information that is highly relevant.
$\text{Conv}_\text{{DwD}}$ effectively has a long-range dependency compared to  $\text{Conv}_\text{{Dw}}$ using dilated convolution.
$\text{Conv}_{\text{1} \times \text{1}}$ captures the relationship between channels in a query or support images. Also, in global attention layer, it can be seen that different heads of the same layer show good performance by simultaneously capturing information from various distances. It is intended to capture both local and global features from multiple heads. But this work shows a similar effect using a single head. This paper overcomes the limitations of not utilizing the local context information of global attention and not considering the relationship between channels.

\begin{figure*}[t]
\centering
\includegraphics[width=0.9\textwidth]{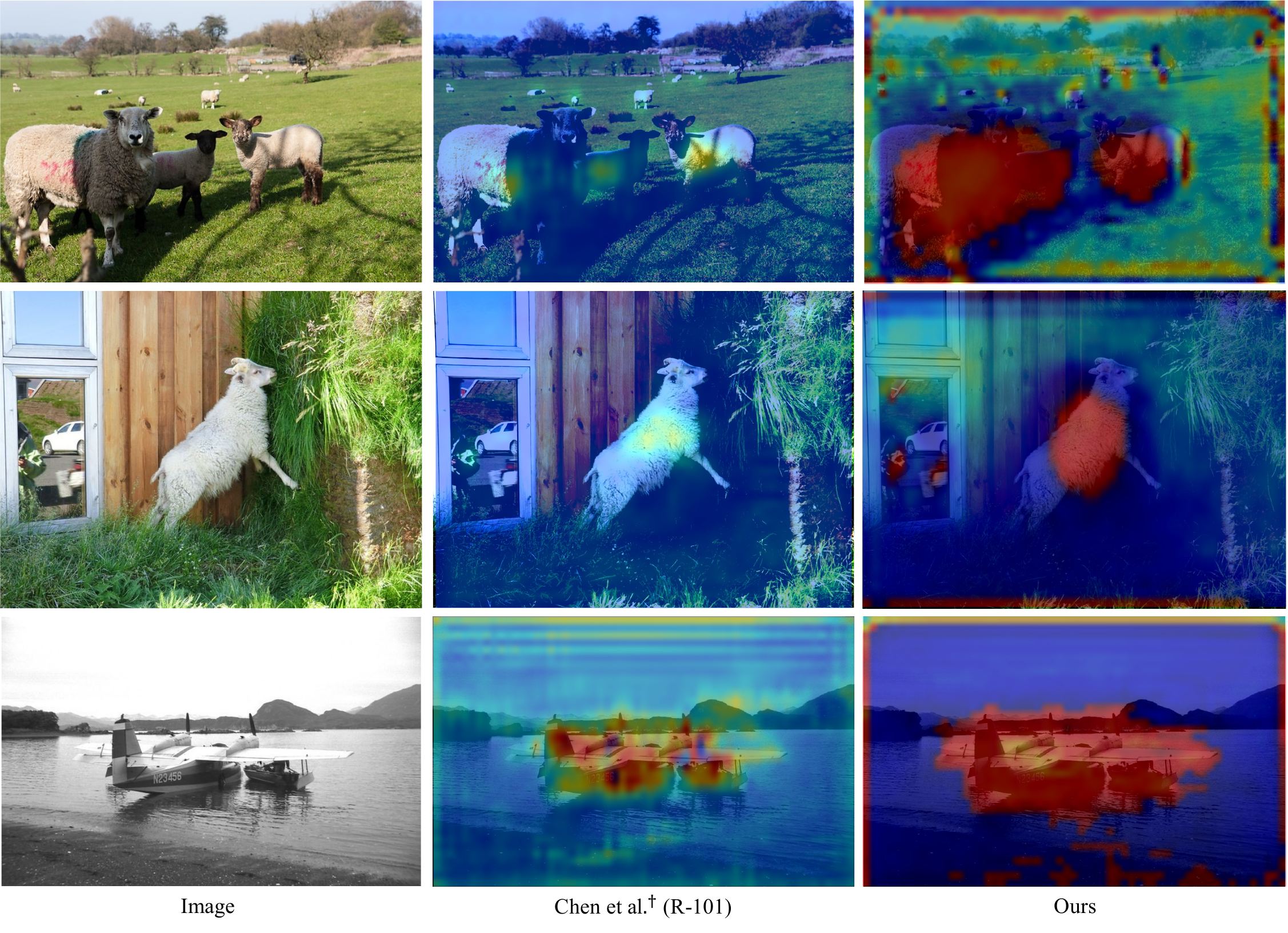} 
\caption{Visualization on the attention map. The first column presents original novel category images. The second column uses only a cross attention between the query and support images and the global attention of the support image. The third column uses additional global attention on the query image and hierarchical attention module.}
\label{fig5}
\end{figure*}

\emph{Then, how can we combine global attention with local attention?}
Since a multi-stage neural network behaves like each model, we design by combining convolution-based attention and global attention for each feature aggregation module.
In Park et al. \cite{park2022vision} experiments, as the depth of the network increased, the variation of the feature map tended to increase. Among them, the convolutional layer makes the variance larger, and the global attention reduces the variance.
In conclusion, global attention ensembles feature map predictions, convolutional layer does not.
Reducing feature map uncertainty helps optimize by ensemble and stabilize the transformed feature map.
So we propose a feature aggregation module that boosts performance by adding convolution-based attentions before global attention.
This also reduces unnecessary degrees of freedom of global attention by using local context information.
Thereby, it allows the model to better capture the pixels that should be focused more on in the global attention layer, which computes its relationship to every pixel.
We introduce a three-stage attention network with a sequentially larger receptive field that utilizes both local information, the relationship between channels, and global information.
As shown in Equations 2 to 7, convolution-based and global attentions are applied to the query and support features and concatenated.
The detailed configuration of HAM can be found in Fig. 3.

\subsection{Meta-Contrastive Learning Module}

\begin{equation}  
\mathcal{L}_{meta}=\sum_{i\in I}\frac{-1}{|M(i)|}\sum_{m\in M(i)}{log(\frac{exp(z_m\cdot z_j/\tau)}{\sum_{a\in A(i)}(exp(z_m \cdot z_a/\tau)})}
\end{equation}
Existing studies have used a meta-learning approach to solve the few-shot learning problem.
However, it is necessary to investigate \emph{whether this really works.}
Meta-learning is a method that helps to evaluate query images well by extracting and aggregating helpful features using support images.
Contrastive learning methods, which are often used in self-supervised learning, are widely used in low-data scenarios and do not show good performance in meta-learning methods.
Matching strategies focus on capturing the inherent variety of query and support images regardless of category rather than on the visual features of each class.
Therefore, binary classification is a more suitable output for modeling the similarity between query and support images than multi-class classification.
Our proposed objective function aims to increase the similarity with the correlation features of the same class and decrease the similarity with the correlation features of different classes if the anchor is a correlation feature of the same class. In Equation 8, let i $\subset$ I $\equiv$ \{1…N\} be the index of a batch image, m $\subset$ M $\equiv$ \{1…M\} be the number of matched correlation features between the query proposal and support features, and j be the index of the other matched correlation feature called \emph{positive} from the same image i. Here, $\tau$ is a scalar temperature parameter, $z_m$ is the anchor, and A(i) $\equiv$ M\textbackslash$\{z_m\}$ is a feature of  \emph{positive} and \emph{negative}, excluding anchor.\\
\indent \emph{Why Meta-CLM make better for meta-learning?} 
If supervised contrastive learning is used to better capture the visual features for each category, the proposed meta-contrastive learning focuses on the correlation between the query image and support image, regardless of the class.
Imagine an extreme example like Fig. 4 at the bottom right.
Assume a situation in which we learn anchor combined with images of the same class \emph{(i.e. cat and cat)} and negative correlation feature of different classes \emph{(i.e. cat and bird)}. It learns to decrease their similarity between anchor and negative correlation feature.
If the model captures only visual features, related features with the same feature map are generated at the same pixel.
However, this module prevents the model from capturing the visual features of each query or support image, as it must learn to decrease the similarity to each other.
Instead, it allows model to extract similarity features to determine whether two images match.

\begin{figure*}[t]
\centering
\includegraphics[width=0.9\textwidth]{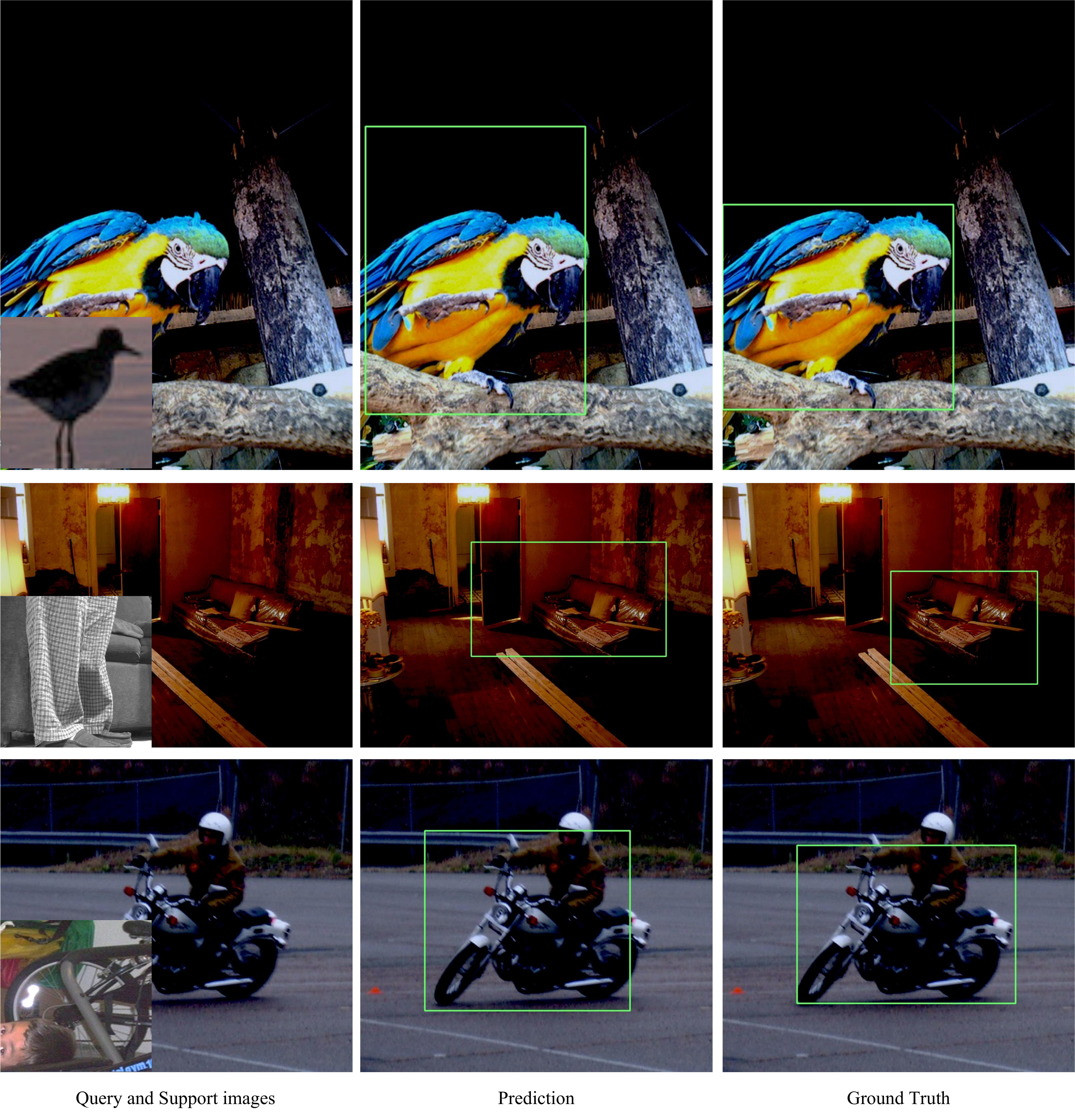} 
\caption{Visualization on qualitative zero-shot results. In the training phase, the model does not learn classes such as bird, motorcycle, and sofa.}
\label{fig6}
\end{figure*}

\begin{table*}[t]
\centering
\caption{Performance on COCO dataset for zero-shot evaluation (1,2,3,5,10, and 30 shots). Avg.: result of averaging 10 multiple runs. -: no result in the paper.}
\includegraphics[width=0.95\textwidth]{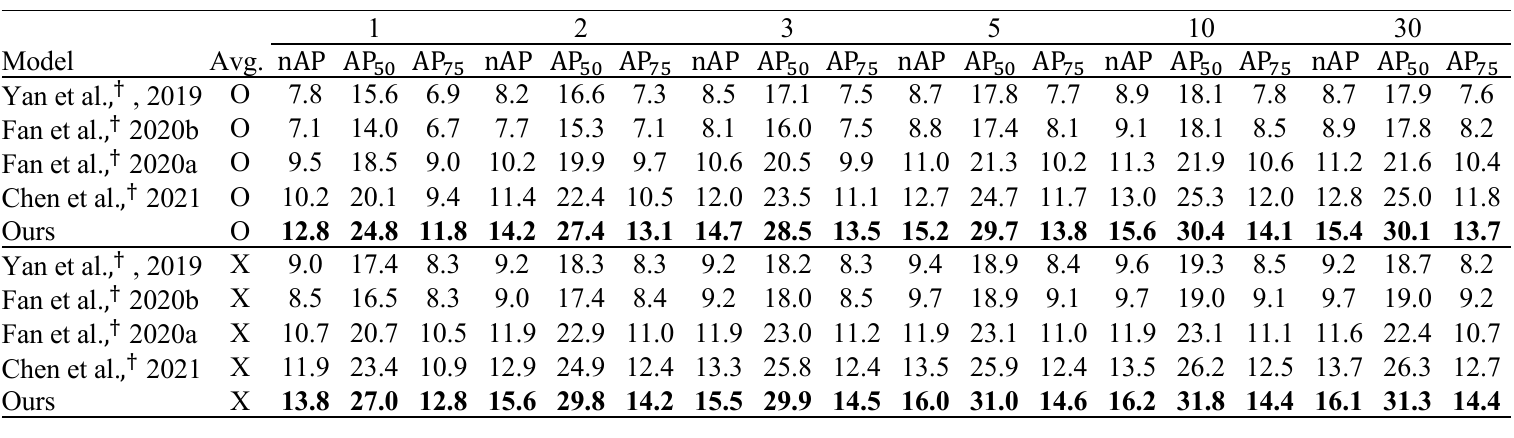} 
\label{table5}
\end{table*}

\subsection{Meta-Contrastive Proposal Feature Encoding}
In the Faster R-CNN pipeline, the RCNN pools cropped the region-of-interest (RoI) features into a fixed size feature map with RoI align, and the feature dimension is mapped to $x \in \Bbb{R}^{7\times7\times d}$. Typically, d = 1024 in the Faster R-CNN pipeline.
The RoI features of the query and support images are embedded in $x\in\Bbb{R}^{7\times 7\times d_{m}}$ through a global average pooling (GAP) and a single fully connected (FC) layer; $d_{m}$ = 128 in this method. Subsequently, we concatenate the query and support features to generate correlation feature and learn to increase the similarity between matched correlation features.

\subsection{Loss Function}

The loss function consists of classification and bounding box regression loss of the RPN, the classification and bounding box regression loss of the RCNN, and includes the loss used in meta-contrastive learning module. The loss function is jointly optimized using Equation 9. The RPN and RCNN loss is the Faster R-CNN \cite{ren2015faster} objective. $\lambda$ of the $\mathcal{L}_{meta}$ was experimentally set to 0.07.

\begin{equation}  
\renewcommand{\arraystretch}{1}
\mathcal{L} =\mathcal{L}_{rpn_{cls}}+\mathcal{L}_{rpn_{reg}}+\mathcal{L}_{rcnn_{cls}}+\mathcal{L}_{rcnn_{reg}}+\lambda \cdot \mathcal{L}_{meta}
\end{equation}

\section{Experiments}
\subsection{Datasets}
Generic FSOD methods use their own datasets and data settings. We conduct the evaluation as a result of 10 multiple runs to make fair comparisons for Pascal VOC and COCO datasets. \\
\textbf{PASCAL VOC}: The dataset consists of 20 classes, divided into 15 base classes and 5 novel classes. We use it for training and evaluation by dividing it into three splits. We use VOC2007 and VOC2012 train/val sets for training, and VOC2007 for evaluation. \\
\textbf{MS-COCO}: We use categories of the PASCAL VOC dataset as novel classes, and the remaining 60 categories as base classes. The validation set is used for testing according to Kang’s split, and the remaining data in the train/val sets are used for training.

\subsection{Experimental Setup}
For a fair comparison, we used a Faster R-CNN framework with ResNet-101 and a feature pyramid network (FPN). We set the learning rate to start at 0.001 and increase by 0.1 times per 10 epochs. Similarly, in the fine-tuning stage, the learning rate decay coefficients were continuously trained from the pre-training stage. We used a stochastic gradient descent (SGD) to optimize the model using a momentum of 0.9 and weight decay of 0.0001. We trained the model using a batch size of 4, and both the pre-training and fine-tuning stages were conducted for 12 epochs on an Nvidia GeForce 3090 GPU.


\subsection{Comparison with State-of-the-Arts}
We compared the performance of the proposed method with that of the latest FSOD methods.
For a fair comparison, the evaluation environment with other models must be the same.
However, unlike other fine-tuning methods, the proposed model does not operate as a loss function for multi-class classification, but operates based on similarity.
Therefore, since the scale of similarity is evaluated without defining different classes, it is difficult for classification to operate efficiently when similar images or objects of different classes are included in the image.
So, in general, it is possible to evaluate the performance of object detection excluding classification in the one-way evaluation method.
Although this defines a somewhat easier task than a model that has to perform both classification and object detection, in a real test environment, an image can contain multiple category objects, and there is no problem in evaluating them in multiple ways.
However, in the current evaluation environment, since one image is viewed and evaluated as having one class, it is inappropriate to evaluate it in multiple ways. So, the evaluation was conducted in a single way, and other meta-learning models were re-implemented and attached.\\
\textbf {Results on Pascal VOC.} 
Table 1 shows the results on PASCAL VOC.
Our proposed method performs better in most novel splits compared to other methods.
It shows good performance in all splits from 1 to 10 shots. In particular, it shows good performance on extremely low-shots or relatively many 10 shots.\\
\textbf {Results on MS COCO.} Our proposed method shows better performance than other methods on all(1,2,3,5,10, and 30) shots.
In particular, our method shows a greater performance improvement at relatively 10 and 30 shots settings. This is because other methods do not fully utilize when averaging a relatively large number of images with large variances, so the performance is rather low.
In Tables 2, 3, and 4, the proposed method improves on the current state-of-the-art by a large margin. In particular, we exceeded SOTA by 14.0, 18.8, 19.8, 20.5, 22.6, and 25.4\% AP for K = 1, 2, 3, 5, 10, and 30 shots on the COCO dataset, respectively. In general, comparing results fairly is difficult because metrics used in previous studies are different. Therefore, we report the results of 10 multiple runs.

\section{Ablation Studies}
\subsection{Qualitative Results}
In this chapter, we demonstrate the effectiveness of our proposed modules via qualitative visualization. Even given few support images, the proposed model is effective in localizing instances. In the training phase, the model does not learn classes such as bird, motorcycle, and sofa. Nevertheless, it is possible to detect objects with only a small number of images. This is a characteristic of metric-based meta-learning, and unlike the fine-tuning method, since it learns based on similarity, it can evaluate query images even for an untrained image if only a support image is given. Figure 6. is the result trained only on the base data $D_{base}$ without any fine-tuning.

\subsection{Hierarchical Attention Visualization}
   This is the result of visualizing the attention map of the existing and proposed methods for a qualitative comparison. Fig. 5 shows the results of evaluating the novel category data for the model trained with novel category data with the fine-tuning step. Compared with the existing method \cite{chen2021dual}, when the attention map is visualized, the weight is more reliably applied to the area where the object is located.

\begin{table}[h]
\centering
\caption{Ablation study results on the effectiveness of each module. For fair comparison, we compare the model with baseline model trained by ResNet-101.}
\includegraphics[width=0.95\columnwidth]{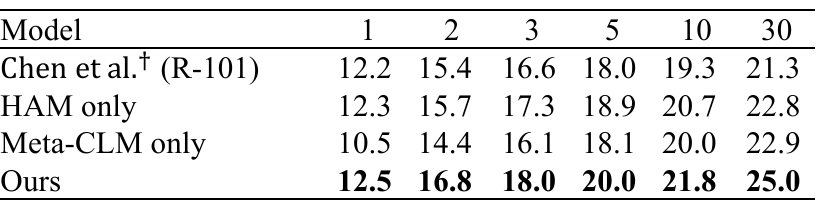} 
\label{table6}
\end{table}

\subsection{Ablation Study on Modules}
We conducted ablation studies to verify the effectiveness of the proposed module. All results were averaged over 10 multiple runs with randomly sampled support datasets according to Kang et al. splits \cite{kang2019few} on COCO.
Table 6 shows the results of the ablation study to analyze whether each module improves in performance. Hierarchical attention can effectively utilize both query and support features. Among methods that obtain the attention score between the query and support features, the model by Hu et al. \cite{hu2021dense} calculates the attention score between the query and support feature, and Chen et al. \cite{chen2021dual} additionally calculates the attention score between the query and support features individually. However, calculating the global attention score of a query is also important. In addition, to compensate for the inability to focus on the local area, which is a disadvantage of global attention, we construct an attention module in a hierarchical structure that extends from local to global, thereby demonstrating significant performance improvements.

\subsection{Ablation Study on Zero-shot Evaluation}
The meta-learning method shows high adaptability even to untrained classes. So, we used the method of evaluating the model trained only with the $D_{base}$ as $D_{novel}$.
All results were averaged over 10 multiple runs with randomly sampled support datasets according to Kang et al. splits on COCO. As the number of shots increased, the training environment and the test environment did not match, so the performance improvement was not large. However, Table 5 shows good performance compared to other methods.

\subsection{Ablation Study on Cross-domain Scenes}
FSOD generally does not assume domain differences, but in a practical evaluation environment (\emph{i.e.} object grasping and autonomous driving), it should work well with unlearned objects or various environments.
So we evaluate the performance when domain or category differences occur.
We tried to verify the results when there is a difference only in the domain and when there is a change in both the category and the domain for fair validation.
Table 7 is the result of training and evaluating with a non-shared category ($C_{train} \cap C_{test}= \emptyset $).
All the models are trained on Pascal VOC 2007 train dataset and tested on COCO 2014 evaluation dataset.
Our proposed method learns domain-agnostic feature and shows better results than the existing method regardless of domain or category.

\begin{table}[h]
\centering
\caption{Performance on COCO dataset for cross-domain evaluation (1,2,3,5,10, and 30 shots).}
\includegraphics[width=0.95\columnwidth]{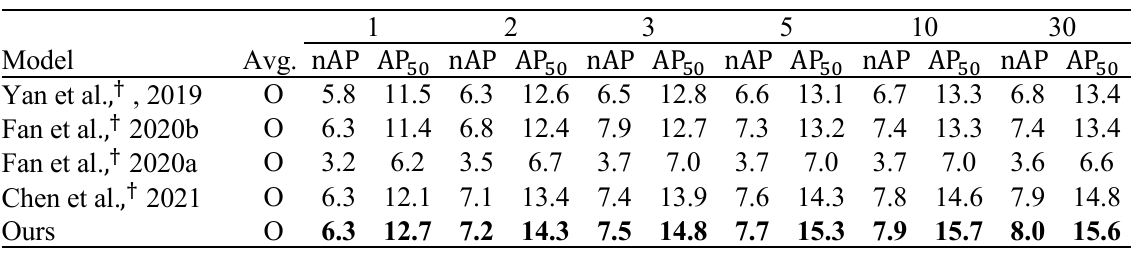}
\label{table7}
\end{table}

\subsection{Ablation Study on Receptive Fields}
Additional experiments were conducted to prove that attention with a hierarchical structure is better than attention with a non- or inverse hierarchical structure. In Table 8, The numbers (\emph{i.e., 5 and 7}) indicate the kernel size of the convolution-based attention. The results of using the attention of the non- and inverse hierarchical structures are shown in rows 4 and 5, respectively. As demonstrated by the results, the method of sequentially expanding the receptive fields from lower to higher layer yields the stable result on COCO.

\begin{table}[h]
\centering
\caption{Comparison of the mAP (mean average precision) of hierarchical, non-hierarchical, and inverse-hierarchical attentions; 14, 21, and 28 indicate the kernel sizes.}
\includegraphics[width=0.95\columnwidth]{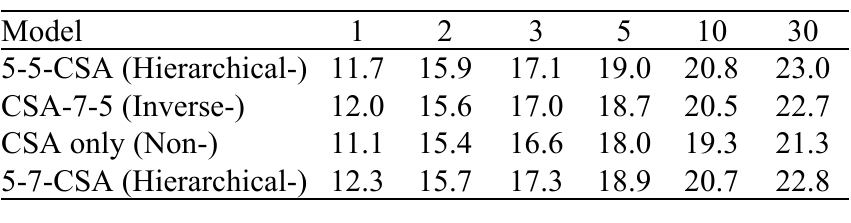}
\label{table8}
\end{table}

\section{Conclusion}
In this study, we propose a hierarchical attention network for FSOD via meta-contrastive learning. In the HAM, we apply a hierarchical attention mechanism that extends from attention that highlights a local area to attention that sequentially explores a global area. Owing to the characteristics of meta-training, the Meta-CLM emphasizes the characteristics of the two correlation features to better extract important characteristics regardless of class. Our proposed method shows a state-of-the-art performance without a fine-tuning stage in low-data scenarios and significant performance improvements compared with existing methods for 10 and 30 shots using fine-tuning on COCO.


%

\ifCLASSOPTIONcaptionsoff
  \newpage
\fi


\bibliographystyle{IEEEtran}
\bibliography{bib.bib}

\begin{thebibliography}{10}
\providecommand{\url}[1]{#1}
\csname url@samestyle\endcsname
\providecommand{\newblock}{\relax}
\providecommand{\bibinfo}[2]{#2}
\providecommand{\BIBentrySTDinterwordspacing}{\spaceskip=0pt\relax}
\providecommand{\BIBentryALTinterwordstretchfactor}{4}
\providecommand{\BIBentryALTinterwordspacing}{\spaceskip=\fontdimen2\font plus
\BIBentryALTinterwordstretchfactor\fontdimen3\font minus
  \fontdimen4\font\relax}
\providecommand{\BIBforeignlanguage}[2]{{%
\expandafter\ifx\csname l@#1\endcsname\relax
\typeout{** WARNING: IEEEtran.bst: No hyphenation pattern has been}%
\typeout{** loaded for the language `#1'. Using the pattern for}%
\typeout{** the default language instead.}%
\else
\language=\csname l@#1\endcsname
\fi
#2}}
\providecommand{\BIBdecl}{\relax}
\BIBdecl

\bibitem{ren2015faster}
S.~Ren, K.~He, R.~Girshick, and J.~Sun, ``Faster r-cnn: Towards real-time
  object detection with region proposal networks,'' \emph{Advances in neural
  information processing systems}, vol.~28, 2015.

\bibitem{lin2017feature}
T.-Y. Lin, P.~Doll{\'a}r, R.~Girshick, K.~He, B.~Hariharan, and S.~Belongie,
  ``Feature pyramid networks for object detection,'' in \emph{Proceedings of
  the IEEE conference on computer vision and pattern recognition}, 2017, pp.
  2117--2125.

\bibitem{he2017mask}
K.~He, G.~Gkioxari, P.~Doll{\'a}r, and R.~Girshick, ``Mask r-cnn,'' in
  \emph{Proceedings of the IEEE international conference on computer vision},
  2017, pp. 2961--2969.

\bibitem{cai2018cascade}
Z.~Cai and N.~Vasconcelos, ``Cascade r-cnn: Delving into high quality object
  detection,'' in \emph{Proceedings of the IEEE conference on computer vision
  and pattern recognition}, 2018, pp. 6154--6162.

\bibitem{hsieh2019one}
T.-I. Hsieh, Y.-C. Lo, H.-T. Chen, and T.-L. Liu, ``One-shot object detection
  with co-attention and co-excitation,'' \emph{Advances in neural information
  processing systems}, vol.~32, 2019.

\bibitem{hu2021dense}
H.~Hu, S.~Bai, A.~Li, J.~Cui, and L.~Wang, ``Dense relation distillation with
  context-aware aggregation for few-shot object detection,'' in
  \emph{Proceedings of the IEEE/CVF conference on computer vision and pattern
  recognition}, 2021, pp. 10\,185--10\,194.

\bibitem{chen2021dual}
T.-I. Chen, Y.-C. Liu, H.-T. Su, Y.-C. Chang, Y.-H. Lin, J.-F. Yeh, W.-C. Chen,
  and W.~Hsu, ``Dual-awareness attention for few-shot object detection,''
  \emph{IEEE Transactions on Multimedia}, 2021.

\bibitem{yan2019meta}
X.~Yan, Z.~Chen, A.~Xu, X.~Wang, X.~Liang, and L.~Lin, ``Meta r-cnn: Towards
  general solver for instance-level low-shot learning,'' in \emph{Proceedings
  of the IEEE/CVF International Conference on Computer Vision}, 2019, pp.
  9577--9586.

\bibitem{raghu2021vision}
M.~Raghu, T.~Unterthiner, S.~Kornblith, C.~Zhang, and A.~Dosovitskiy, ``Do
  vision transformers see like convolutional neural networks?'' \emph{Advances
  in Neural Information Processing Systems}, vol.~34, pp. 12\,116--12\,128,
  2021.

\bibitem{park2022vision}
N.~Park and S.~Kim, ``How do vision transformers work?'' \emph{arXiv preprint
  arXiv:2202.06709}, 2022.

\bibitem{vinyals2016matching}
O.~Vinyals, C.~Blundell, T.~Lillicrap, D.~Wierstra \emph{et~al.}, ``Matching
  networks for one shot learning,'' \emph{Advances in neural information
  processing systems}, vol.~29, 2016.

\bibitem{sung2018learning}
F.~Sung, Y.~Yang, L.~Zhang, T.~Xiang, P.~H. Torr, and T.~M. Hospedales,
  ``Learning to compare: Relation network for few-shot learning,'' in
  \emph{Proceedings of the IEEE conference on computer vision and pattern
  recognition}, 2018, pp. 1199--1208.

\bibitem{xiao2020few}
Y.~Xiao and R.~Marlet, ``Few-shot object detection and viewpoint estimation for
  objects in the wild,'' in \emph{European conference on computer
  vision}.\hskip 1em plus 0.5em minus 0.4em\relax Springer, 2020, pp. 192--210.

\bibitem{zhang2021accurate}
L.~Zhang, S.~Zhou, J.~Guan, and J.~Zhang, ``Accurate few-shot object detection
  with support-query mutual guidance and hybrid loss,'' in \emph{Proceedings of
  the IEEE/CVF Conference on Computer Vision and Pattern Recognition}, 2021,
  pp. 14\,424--14\,432.

\bibitem{zhang2021meta}
G.~Zhang, Z.~Luo, K.~Cui, and S.~Lu, ``Meta-detr: Image-level few-shot object
  detection with inter-class correlation exploitation,'' \emph{arXiv preprint
  arXiv:2103.11731}, 2021.

\bibitem{li2021transformation}
A.~Li and Z.~Li, ``Transformation invariant few-shot object detection,'' in
  \emph{Proceedings of the IEEE/CVF Conference on Computer Vision and Pattern
  Recognition}, 2021, pp. 3094--3102.

\bibitem{lin2014microsoft}
T.-Y. Lin, M.~Maire, S.~Belongie, J.~Hays, P.~Perona, D.~Ramanan,
  P.~Doll{\'a}r, and C.~L. Zitnick, ``Microsoft coco: Common objects in
  context,'' in \emph{European conference on computer vision}.\hskip 1em plus
  0.5em minus 0.4em\relax Springer, 2014, pp. 740--755.

\bibitem{kang2019few}
B.~Kang, Z.~Liu, X.~Wang, F.~Yu, J.~Feng, and T.~Darrell, ``Few-shot object
  detection via feature reweighting,'' in \emph{Proceedings of the IEEE/CVF
  International Conference on Computer Vision}, 2019, pp. 8420--8429.

\bibitem{fan2020few}
Q.~Fan, W.~Zhuo, C.-K. Tang, and Y.-W. Tai, ``Few-shot object detection with
  attention-rpn and multi-relation detector,'' in \emph{Proceedings of the
  IEEE/CVF Conference on Computer Vision and Pattern Recognition}, 2020, pp.
  4013--4022.

\bibitem{qiao2021defrcn}
L.~Qiao, Y.~Zhao, Z.~Li, X.~Qiu, J.~Wu, and C.~Zhang, ``Defrcn: Decoupled
  faster r-cnn for few-shot object detection,'' in \emph{Proceedings of the
  IEEE/CVF International Conference on Computer Vision}, 2021, pp. 8681--8690.

\bibitem{cao2021few}
Y.~Cao, J.~Wang, Y.~Jin, T.~Wu, K.~Chen, Z.~Liu, and D.~Lin, ``Few-shot object
  detection via association and discrimination,'' \emph{Advances in Neural
  Information Processing Systems}, vol.~34, pp. 16\,570--16\,581, 2021.

\bibitem{sun2021fsce}
B.~Sun, B.~Li, S.~Cai, Y.~Yuan, and C.~Zhang, ``Fsce: Few-shot object detection
  via contrastive proposal encoding,'' in \emph{Proceedings of the IEEE/CVF
  Conference on Computer Vision and Pattern Recognition}, 2021, pp. 7352--7362.

\bibitem{song2022comprehensive}
Y.~Song, T.~Wang, S.~K. Mondal, and J.~P. Sahoo, ``A comprehensive survey of
  few-shot learning: Evolution, applications, challenges, and opportunities,''
  \emph{arXiv preprint arXiv:2205.06743}, 2022.

\bibitem{antonelli2022few}
S.~Antonelli, D.~Avola, L.~Cinque, D.~Crisostomi, G.~L. Foresti, F.~Galasso,
  M.~R. Marini, A.~Mecca, and D.~Pannone, ``Few-shot object detection: A
  survey,'' \emph{ACM Computing Surveys (CSUR)}, vol.~54, no. 11s, pp. 1--37,
  2022.

\bibitem{han2022few}
G.~Han, J.~Ma, S.~Huang, L.~Chen, and S.-F. Chang, ``Few-shot object detection
  with fully cross-transformer,'' in \emph{Proceedings of the IEEE/CVF
  Conference on Computer Vision and Pattern Recognition}, 2022, pp. 5321--5330.

\bibitem{han2022meta}
G.~Han, S.~Huang, J.~Ma, Y.~He, and S.-F. Chang, ``Meta faster r-cnn: Towards
  accurate few-shot object detection with attentive feature alignment,'' in
  \emph{Proceedings of the AAAI Conference on Artificial Intelligence},
  vol.~36, no.~1, 2022, pp. 780--789.

\bibitem{guo2022visual}
M.-H. Guo, C.-Z. Lu, Z.-N. Liu, M.-M. Cheng, and S.-M. Hu, ``Visual attention
  network,'' \emph{arXiv preprint arXiv:2202.09741}, 2022.

\bibitem{li2021beyond}
B.~Li, B.~Yang, C.~Liu, F.~Liu, R.~Ji, and Q.~Ye, ``Beyond max-margin: Class
  margin equilibrium for few-shot object detection,'' in \emph{Proceedings of
  the IEEE/CVF Conference on Computer Vision and Pattern Recognition}, 2021,
  pp. 7363--7372.

\bibitem{li2021libfewshot}
W.~Li, C.~Dong, P.~Tian, T.~Qin, X.~Yang, Z.~Wang, J.~Huo, Y.~Shi, L.~Wang,
  Y.~Gao \emph{et~al.}, ``Libfewshot: A comprehensive library for few-shot
  learning,'' \emph{arXiv preprint arXiv:2109.04898}, 2021.

\bibitem{hospedales2021meta}
T.~Hospedales, A.~Antoniou, P.~Micaelli, and A.~Storkey, ``Meta-learning in
  neural networks: A survey,'' \emph{IEEE transactions on pattern analysis and
  machine intelligence}, vol.~44, no.~9, pp. 5149--5169, 2021.

\bibitem{han2021query}
G.~Han, Y.~He, S.~Huang, J.~Ma, and S.-F. Chang, ``Query adaptive few-shot
  object detection with heterogeneous graph convolutional networks,'' in
  \emph{Proceedings of the IEEE/CVF International Conference on Computer
  Vision}, 2021, pp. 3263--3272.

\bibitem{chen2020simple}
T.~Chen, S.~Kornblith, M.~Norouzi, and G.~Hinton, ``A simple framework for
  contrastive learning of visual representations,'' in \emph{International
  conference on machine learning}.\hskip 1em plus 0.5em minus 0.4em\relax PMLR,
  2020, pp. 1597--1607.

\bibitem{zhang2020cooperating}
W.~Zhang, Y.-X. Wang, and D.~A. Forsyth, ``Cooperating rpn's improve few-shot
  object detection,'' \emph{arXiv preprint arXiv:2011.10142}, 2020.

\bibitem{fan2020fgn}
Z.~Fan, J.-G. Yu, Z.~Liang, J.~Ou, C.~Gao, G.-S. Xia, and Y.~Li, ``Fgn: Fully
  guided network for few-shot instance segmentation,'' in \emph{Proceedings of
  the IEEE/CVF conference on computer vision and pattern recognition}, 2020,
  pp. 9172--9181.

\bibitem{wang2020frustratingly}
X.~Wang, T.~E. Huang, T.~Darrell, J.~E. Gonzalez, and F.~Yu, ``Frustratingly
  simple few-shot object detection,'' \emph{arXiv preprint arXiv:2003.06957},
  2020.

\bibitem{li2020meta}
S.~Li, W.~Song, S.~Li, A.~Hao, and H.~Qin, ``Meta-retinanet for few-shot object
  detection.'' in \emph{BMVC}, 2020.

\bibitem{wu2020multi}
J.~Wu, S.~Liu, D.~Huang, and Y.~Wang, ``Multi-scale positive sample refinement
  for few-shot object detection,'' in \emph{European conference on computer
  vision}.\hskip 1em plus 0.5em minus 0.4em\relax Springer, 2020, pp. 456--472.

\bibitem{khosla2020supervised}
P.~Khosla, P.~Teterwak, C.~Wang, A.~Sarna, Y.~Tian, P.~Isola, A.~Maschinot,
  C.~Liu, and D.~Krishnan, ``Supervised contrastive learning,'' \emph{Advances
  in Neural Information Processing Systems}, vol.~33, pp. 18\,661--18\,673,
  2020.

\bibitem{wang2019meta}
Y.-X. Wang, D.~Ramanan, and M.~Hebert, ``Meta-learning to detect rare
  objects,'' in \emph{Proceedings of the IEEE/CVF International Conference on
  Computer Vision}, 2019, pp. 9925--9934.

\bibitem{vaswani2017attention}
A.~Vaswani, N.~Shazeer, N.~Parmar, J.~Uszkoreit, L.~Jones, A.~N. Gomez,
  {\L}.~Kaiser, and I.~Polosukhin, ``Attention is all you need,''
  \emph{Advances in neural information processing systems}, vol.~30, 2017.

\bibitem{snell2017prototypical}
J.~Snell, K.~Swersky, and R.~Zemel, ``Prototypical networks for few-shot
  learning,'' \emph{Advances in neural information processing systems},
  vol.~30, 2017.

\bibitem{he2016deep}
K.~He, X.~Zhang, S.~Ren, and J.~Sun, ``Deep residual learning for image
  recognition,'' in \emph{Proceedings of the IEEE conference on computer vision
  and pattern recognition}, 2016, pp. 770--778.

\bibitem{everingham2010pascal}
M.~Everingham, L.~Van~Gool, C.~K. Williams, J.~Winn, and A.~Zisserman, ``The
  pascal visual object classes (voc) challenge,'' \emph{International journal
  of computer vision}, vol.~88, no.~2, pp. 303--338, 2010.

\end{thebibliography}

%




\end{document}